# Deep Convolutional Networks as Models of Generalization and Blending Within Visual Creativity

Graeme McCaig
Simon Fraser University,
Canada
gmccaig@sfu.ca

Steve DiPaola
Simon Fraser University,
Canada
sdipaola@sfu.ca

Liane Gabora
University of British Columbia,
Canada
liane.gabora@ubc.ca

## Abstract

We examine two recent artificial intelligence (AI) based deep learning algorithms for visual blending in convolutional neural networks (Mordvintsev et al. 2015, Gatys et al. 2015). To investigate the potential value of these algorithms as tools for computational creativity research, we explain and schematize the essential aspects of the algorithms' operation and give visual examples of their output. We discuss the relationship of the two algorithms to human cognitive science theories of creativity such as conceptual blending theory and honing theory, and characterize the algorithms with respect to generation of novelty and aesthetic quality.

## Introduction

It has been suggested for some time that neural networks are appealing models for aspects of creativity such as the ability to blend concepts based on their wider context and associations (Boden 2004). The recent deep learning trend (Bengio et al. 2013) utilizes multi-layer artificial neural networks, is inspired in part by theorized principles of human brain function, and has produced impressive practical results in visual processing. In particular, Deep Dream (Mordvintsev et al. 2015) and Neural Style Transfer (Gatys et al. 2015) are algorithms, based on deep-learning convolutional neural networks (CNNs) (Krizhevsky et al. 2012), that blend visual qualities from multiple source images to create a new output image. (We adopt the shortened name "Deep Style" for Neural Style Transfer—Deep Dream and Deep Style are then abbreviated DD and DS.) Since their introduction, DD and DS have evoked public attention and speculation about their level of creativity and their ability to replace human artists. Figure 1 provides an example of our results combining DD with additional painterly (stroke-based) rendering.

Here we consider how DD/DS and related CNN-based algorithms can play a role in computational creativity research as cognitively inspired mechanisms for visual blending and imagination. Both computer cognitive modeling research and results-focused computational creative systems (e.g. generative art systems) could be targets for DD/DS type algorithms. First, we examine how DD/DS fit into certain existing cognitive theories of creativity.

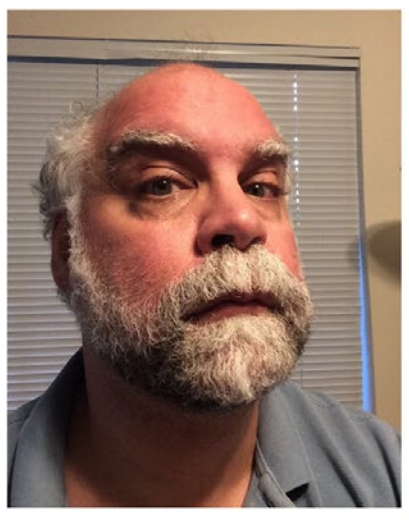

a.

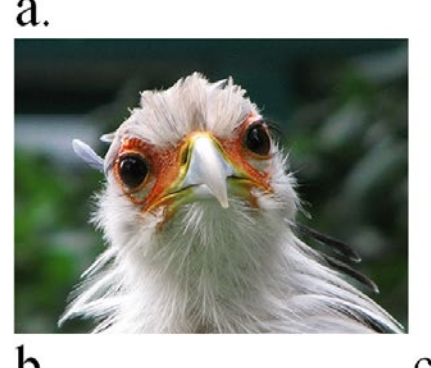

b.

c.

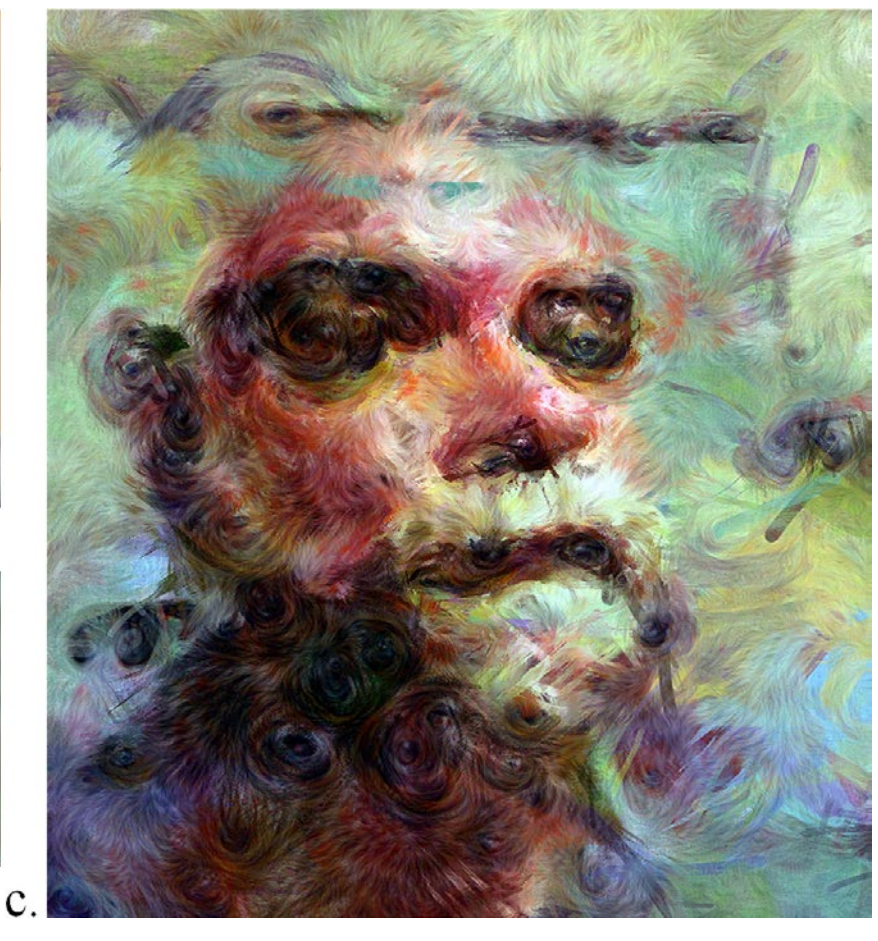

Figure 1. Using Deep Dream, the bird image (b) is used as a guide with source (a), thereby conceptually blending the two (c).

## Visual Concept Blending and Honing Theory

*Conceptual blending* (Fauconnier & Turner 1998) is a proposed cognitive mechanism in which new concepts are obtained by integrating multiple pre-existing conceptual spaces. It has been suggested as an important mechanism underlying human creativity (Pereira & Cardoso 2002). Existing approaches to computational modeling of conceptual blending include symbolic AI-based systems (Besold & Plaza 2015) and the approach of Thagard & Stewart (2011) which combines neural patterns using a convolution operation (different from convolution as used in CNNs). Another recent direction in modeling conceptual blending uses a mathematical formalism based on quantum mechanics (Aerts & Gabora 2016). While the contextuality and noncompositionality of concepts makes them resistant to mathematical description, quantum formalism provides a natural spatial representation in which variables are natively context specific.

Theory and modeling of conceptual blending often includes *visual blends*, in which the novel output of a blending operation is expressed as an image. Visual blending is often framed in terms of combining high-level, verbalizable concepts (Xiao & Linkola 2015; Martins et al. 2015; Confalonieri et al. 2015). However, the strength of DD/DS lies in their ability to combine lower-level yet abstract im-age qualities such as shapes, textures and arrangements, extracting such qualities from input images without explicit labelling. Lower-level image-quality blending is likely a component of visual *imagination* (Richardson 2015; Heath et al. 2015), and fits the idea of preconceptual creativity discussed in (Takala 2015). Thus, we suggest that DD/DS are promising as computational tools for exploring preconceptual visual blending and imagination, which are relevant to creativity and art creation.

Another theoretical framework for cognitive creativity is honing theory (Gabora 2005). Honing theory predicts that creativity involves the merging and interference of memory items resulting in a single cognitive structure that is *ill-defined*, and can be said to exist in a state of potentiality, and which can be formally described as a superposition state. The idea becomes increasingly well-defined, and transforms from potential to actual through interaction with internally and externally generated contexts. The idea could actualize in different ways depending on the contexts the idea interacts with, or perspectives it is viewed from. Innovative concept combinations are thought to take place in an *associative* mode of thought, in which more features of the object of thought are held in mind, and thus there are more associative routes available to items that have previously gone unnoticed (Gabora & Ranjan, 2013). The transition between a tightly focused *analytic* mode of thought and a widely focused associative mode of thought typically takes place many times in the creative process, and facility in making this transition known as *contextual focus* is important to creative ability (DiPaola and Gabora 2009).

As we will discuss and illustrate, DD and DS share qualities with mechanisms proposed in honing theory such as:

- A repeated dual-phase search process in which images are analyzed from low-level to a collection of abstract, higher-level perspectives (features) – resemblance or emphasis is found at this higher level, and then transferred back to the pixel level
- The interaction of visual concepts in working memory with a previously learned network of visual features built up over time
- Identifying points of resonance or visual metaphors across diverse images based on points of similarity in abstract feature association space

## Deep Learning Convolutional Neural Nets

Deep learning is a collection of network-based machine learning methods that are notable for their power and breakthrough performance in tasks such as object recognition (Krizhevsky et al. 2012), as well as their similarities to aspects of human vision and brain function (DiCarlo et al. 2012). When network training is complete, running novel input data through the network models perception, and some network types are capable of using feedback connections to create novel data generalized from what has been learned (Salakhutdinov and Hinton 2009). Deep generative models have been proposed as useful for cognitive modelling (Zorzi et al. 2013). A deep belief net is used as a perception module in Spaun Cognitive Architecture (Stewart et al. 2012), and a deep Boltzmann machine used as a model for Charles Bonnet Syndrome (hallucination) by Reichert et al. (2013).

Machine learning has traditionally depended on hand-engineering of features. Recent techniques focus on representation learning, which strives to automatically "extract and organize relevant information from the data", a step toward human-like AI (Bengio et al 2013). Deep learning techniques, which compose multiple non-linear transformations of the data, have become an important focus in representation learning. Much of the power of deep networks comes from their ability to naturally represent abstraction. Abstraction occurs when a certain symbol or encoding element stands for a broader range of specific instantiations. The nested, hierarchical structure of abstraction maps directly to the connection structure of a deep net.

The convolutional neural network (CNN) (LeCun et al. 1998) is a deep feedforward neural net architecture usually trained with backpropagation. The CNN architecture aids generalization, efficient training, and invariance to input distortions by incorporating reasonable assumptions about the input image domain through mechanisms of local receptive fields, shared weights, and sub-sampling. "Core object recognition" in the human brain is thought (DiCarlo et al. 2012) to use an alternating structure of selectivity and tolerance transformations, similar to convolutional nets.

## Visual Blend CNNs: Deep Dream & Style

Using our code and scripts based on these algorithms as well as a simple selfie source of one of the authors we explore and explain the principles of Deep Dream (Mordyintsev et al. 2015), focusing on guide-image mode, and Deep Style (Gatys et al 2015). DD was developed by Google researchers and introduced via a blog-post and open source code (github.com/google/deepdream). The purposes of DD are not only to "check what [a] network learned during training" but also to provide "a new way to remix visual concepts—or perhaps even shed a little light on the roots of the creative process in general" (Mordyintsev et al 2015). DS grew out of research in texture-transfer and texture-synthesis—the authors view the algorithm as a way to separate out the "content" from one image and the "style" from another, fusing them in a novel image. These two algorithms both use pre-trained CNNs to generate a transformed version of a source image, emphasizing certain semantic and/or stylistic qualities.

The two algorithms are similar in many respects, as Figure 2 illustrates. Each begins with a style-source image, which is propagated from the lowest (pixel) layer to a selected set of higher layer(s)—the higher layers give a more generalized encoding of the image in terms of abstract features. This encoding is stored as a guide-style tensor. In DS, the guide-style tensor is accompanied by a guide-content tensor, found by propagating the content-source image through the network (to a layer higher than that of the guide tensor). The lack of guide-content tensor in DD means that DD is more divergent, wandering farther from the source image over successive iterations into patterns and associations that depend more on the network's learned biases and less on what was originally present.

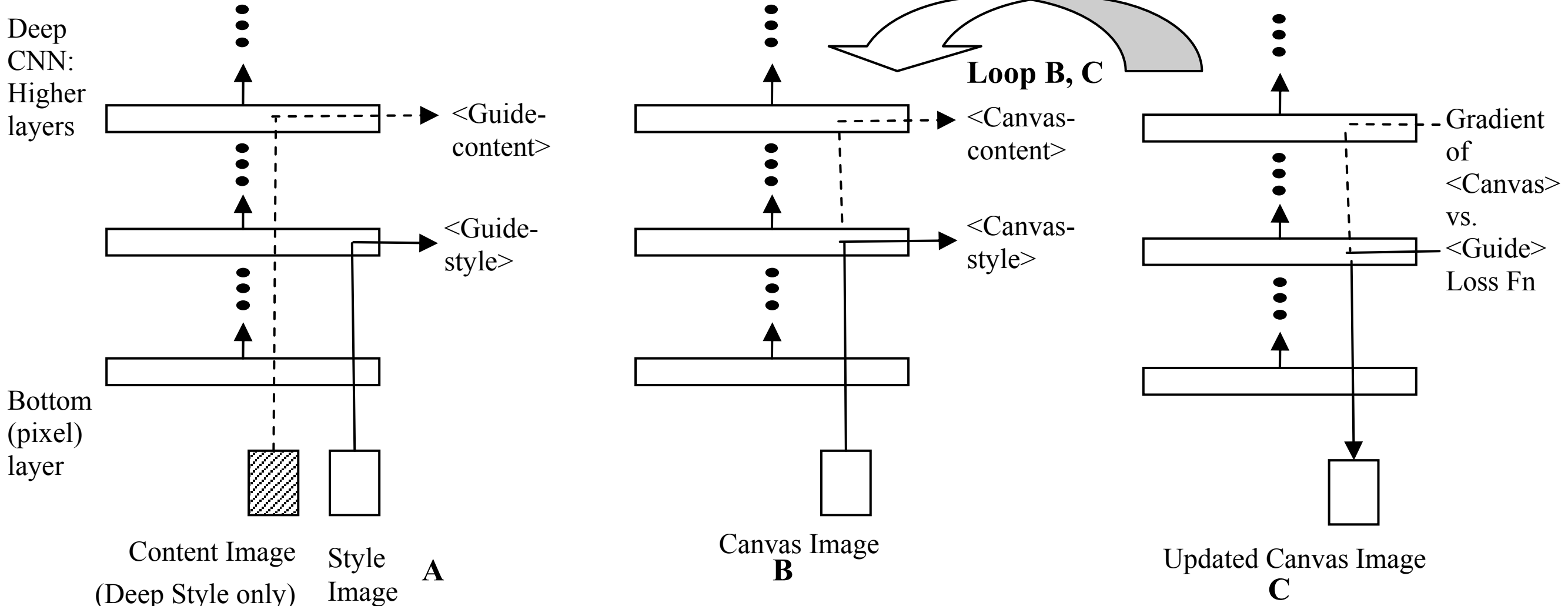

Figure 2. Schema illustrating the operation of Deep Dream and Deep Style algorithms. Dashed lines apply only to Deep Style.

At this point, the algorithms initialize the set of pixels that will be gradually transformed into the output image (we call this set the canvas image due to its dynamically updated nature). In DD, the canvas image is set equal to the content-source image, while in DS the canvas image is initialized as random noise. The canvas image is propagated up through the network to the same layers as the guide-style tensor and guide-content tensor, respectively, yielding the canvas-style tensor and the canvas-content tensor.

A loss function is defined to measure the similarity of the canvas-style tensor vs. guide-style tensor (and also canvas-content tensor vs. guide-content tensor for DS). From this loss function a gradient is found, indicating how the canvas tensors should be incrementally changed to be more similar to the guide tensors. This gradient is back-propagated to the lowest layer, and a small step in the desired direction is applied as a change to pixel values. Repeating the cycle of (update pixels, propagate forward, find gradient, propagate downward) constitutes a gradient ascent process.

The nature of the guide tensors and loss functions determines which visual attributes are drawn from the source and guide images in a blend. DD maximizes a dot product between feature vectors from the canvas activation and the feature vectors taken from *best-matching locations* in the guide activation. Thus a DD image will tend to pick up on certain types of shape and texture, found within the guide image, that bear similarity to its own shapes and textures.

In DS, the guide-style tensors are computed as Gram matrices, a set of correlations among the features within each layer. Compared to the dot-product method this approach seems to more faithfully capture the look of specific shapes and textures from the guide image, particularly with respect to color schemes. Some generalization does take place, allowing regions of the canvas to be *similar* to the guide rather than slavishly copying patches. This use of Gram matrices also encourages the canvas as a whole to capture different aspects of the guide image rather than fixate on a particular visual component as DD often does. DS combines the Gram matrix approach with a simpler vector-similarity measure used for the content-guide, ensuring that the recognizable content (spatial layout and high-level features) of the source image is maintained.

The two algorithms can be applied to various CNN architectures and training sets. In this paper we employ GoogLeNet (Szegedy et al. 2015) for DD and VGG (Simonyan & Zisserman 2014) for DS, as did the algorithms' designers. The network weights come from training on the ImageNet dataset (except in the experiment illustrated in Figure 3f, using a network trained on the Comprehensive Cars dataset). Our implementations of DD and DS are built on the Caffe CNN library (Jia et al. 2014).

Deep Dream can alternatively be run in a no-guide-image mode (what we might call "free hallucination"). In this mode, the quantity optimized is the L2-norm of activation for one selected network layer (meaning whichever of that layer's nodes are most strongly activated will tend to become more so).

## Generalization and Two-Phase Aspects

As Figure 2 helped to illustrate, DD and DS incorporate mechanisms proposed to play an important role in creative cognition. Crucially, to create a blend of two images, these algorithms first generalize each image by propagating it through a deep CNN and representing it according to the resulting tensor encoding at a certain network layer(s). Depending on the height and type of the network layer, the given encoding analyzes the image according to a particular set of visual features. Much as in the parable of the blind men who each describe an elephant in a different way, different layers "see the image" in different ways, the visual blend depends on points of similarity between guide and source, as viewed from the perspective of a certain network layer encoding. In turn, the nature of a layer en-coding depends on both the network architecture and the original training data, which caused features to be developed as a form of long-term memory.

To enhance similarity between the source image and the guide, the algorithms use a reiterative two-phase creative process of alternating divergence and convergence; similarities found at a high/abstract level are manifested back at the pixel level.

## Algorithm Input/Output Examples

This section provides examples of images processed by the DD and DS algorithms, and discusses them in terms of creative cognitive mechanisms.

Figure 3 presents results obtained using Deep Dream in no-guide-image mode. This illustrates the type of visual features according to which different network layers encode and interpret the image. We note that such visualization does not capture the entire range of visual shapes/textures potentially encoded by a given layer. Rather it tends to display a kind of layer bias—a strong regime of activation (in the L2-norm sense) into which the layer tends to “find its way” using gradient ascent search.

Certain characteristics of the output of Deep Dream (guided or not) as well as Deep Style are particularly evident in Figure 3. Firstly, the algorithm does not merely superimpose layer-specific features in a random way over the image; rather features tend to be emphasized and grown starting from those image regions that already contain said features. For example, a layer that emphasizes circle and arc shapes tends to place them in pre-existing arc-shaped parts of the image, such as the curved orbital region around a subject’s eye. On the other hand, if the algorithm is run for many iterations, all image regions will eventually be forced in the direction of a high-activating feature, essentially making something out of nothing.

Another striking aspect of the DD output is that the textures and shapes convey a sense of completion and good continuation, or flow. For example, in Figure 3b, we see a repeated plate or petal pattern in which most of the plates are similar size and shape, not overlapping or being interrupted. This arises from the type of optimization performed by the search process; the total activation of a layer is more enhanced when neighboring features work together without overlapping or disrupting each other.

Two notable “clichéd” aspects of DD deserve comment:

1. Being designed as a bottom-up discrimination network, GoogLeNet discards much of the data about tonic color of regions, but retains color-contrast near edges. When feature detection is optimized, contrasting color bands tend to be created while losing much of the source color information (color remains relevant for objects such as fire trucks where it is an important identity cue).

2. The ImageNet training data contains a bias towards animal types as a large portion of the 1k labelled categories, with a particular emphasis on fine-grained distinction of dog breeds. Hence, a large portion of the network’s capacity has attuned to the task of detecting dog faces and parts. This creates a bias in the encoding space wherein shapes and patterns are likely to be treated as relevant to dog features; hence dog features emerge. Figure 3f shows results obtained using a network trained on car images (CompCars database); indeed, car features emerge.

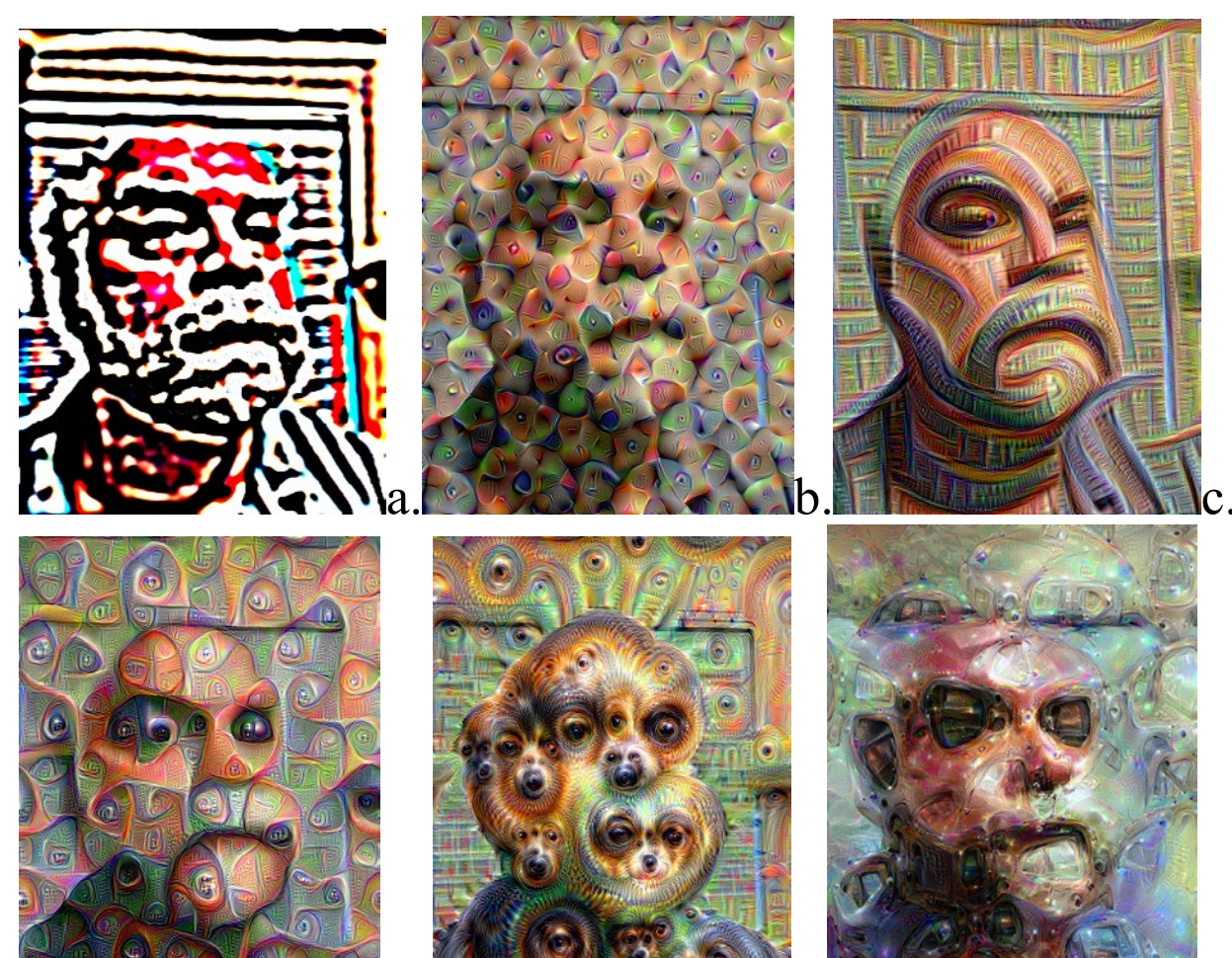

Figure 3. Comparing no-guide output of Deep Dream for low to high network layers (a-e), and alternate training set (f).

We now examine results from our DD implementation when a guide image is used, exploring different guide images and sources along with different layers used as target for guide- and canvas- tensors. Many of our results use as input a simple selfie portrait from a front facing iPad (800x600 pixel resolution) in office lighting to show these results from our systems can be obtained from simple non-professional source material.

Figure 4 provides a sample set of input/output images, showing how the algorithm generalizes visual features depending on the network layer. In 4b, butterfly features are represented as grid-like patterns and black/orange/yellow colors, while in 4c more complete shapes such as wings and limbs/antennae emerge. These particular shapes are not directly visible in the butterfly image nor in the face image, but emerge when generalizing from both. In Figures 4d and 4e (obtained using a different source with the same guide) the wings vary in shape depending on the source image as well. These results demonstrate that the wing shape emerges from the particular combination of input images without being explicitly present in either.

Figure 5 explores the difference in source vs. guide image roles by using a certain two images in alternating roles. Attributes generalized from the fire truck include rectangular and curved shapes (and a reddish tinge) at the lower level, while many truck-like components emerge at a higher level. The features abstracted from the face seem to be focused on hair-like patterns, as well as the transformation of the truck’s windows into orbital region- or eye-like curves. Figure 6 shows the different effect of DS: DS captures more specific-looking fragments from the guide image (giving a collaged look), more faithful colors, and a more balanced distribution of colors and shapes from across all parts of the guide. Figure 7 shows two guide/source combinations that produced unexpected results, illustrating how the learned background knowledge of the network plays a key role in how guide and source image features are interpreted and hence generalized/reapplied. In Figure 7b it is likely the network has not been specifically trained on many angular-origami-pattern images, yet it "understands" the guide image in terms of a range of patterns and particularly crease-like dark diagonal lines. In Fig 7d, the network features do a good job of representing several bee-like aspects: yellowness, fuzziness, dark eyes and dangling legs. Yet, the algorithm also falls prey to the aforementioned training set bias of seeing dog faces, creating a fanciful 3-way dog/bee/face hybrid. Due to the fact that DS adheres closely to the look of specific guide image fragments, it is quite successful at painting style transfer, i.e. applying the colorist and brush stroking style from a specific painting to a new source image to imitate, e.g., Rembrandt. Figure 8 shows two examples.

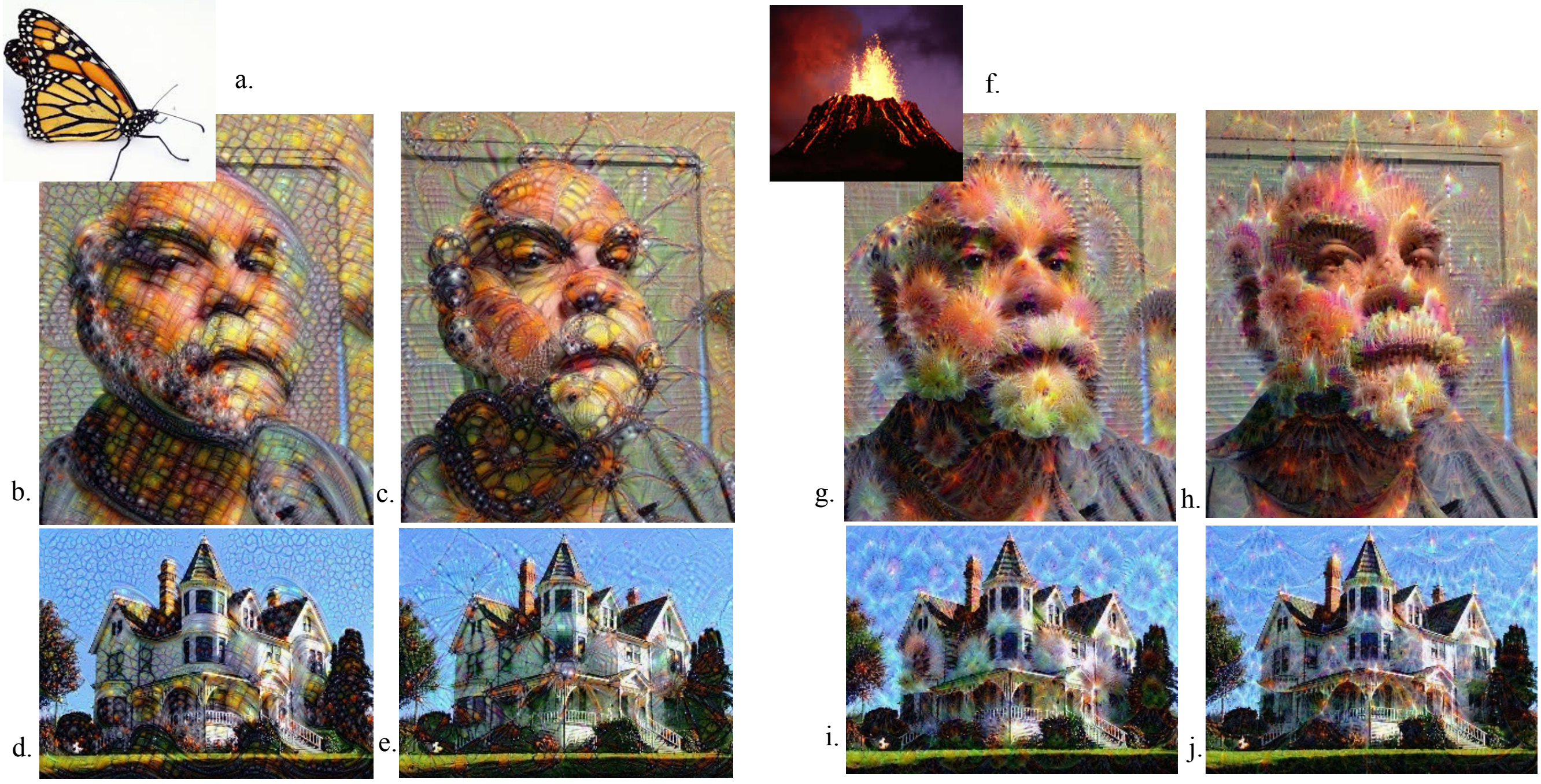

Figure 4. Deep Dream examples using different source and guide images and net layers. Images (b-e) use guide (a), images (g-j) use guide (f). Images (b,d,g,i) use a lower net-layer parameter while (c,e,h,j) use a higher net-layer parameter. Butterfly image by Ano Lobb (Flickr, CC BY 2.0). Original house image by Jan Tik (Flickr, CC BY 2.0). Volcano image courtesy of US Geological Survey.

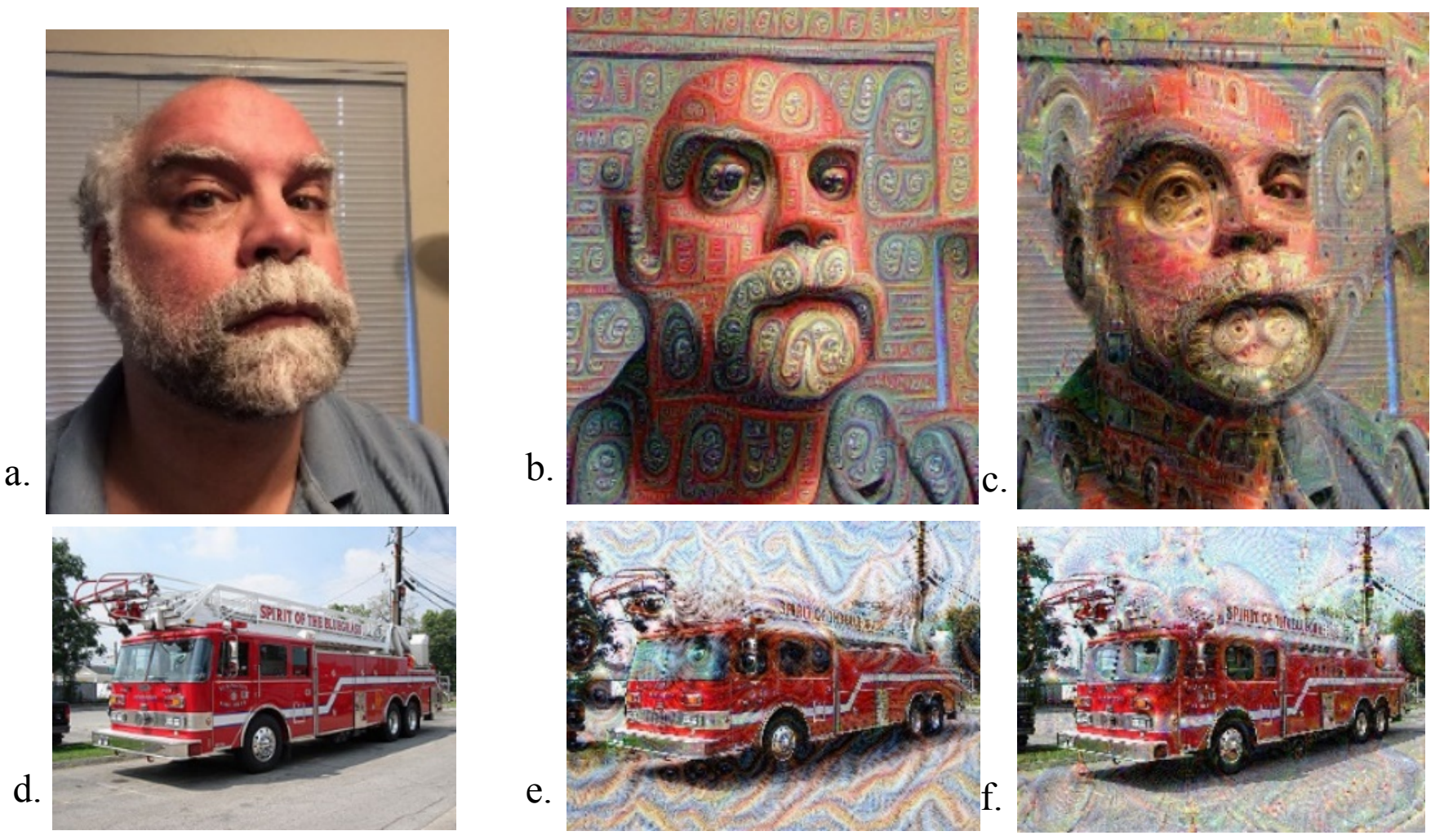

Figure 5. Deep Dream. Top row: face (a) as source with fire engine (d) guide; bottom row: fire engine source with face guide. (b,e) use lower net-layer, (c,f) use higher net-layer.

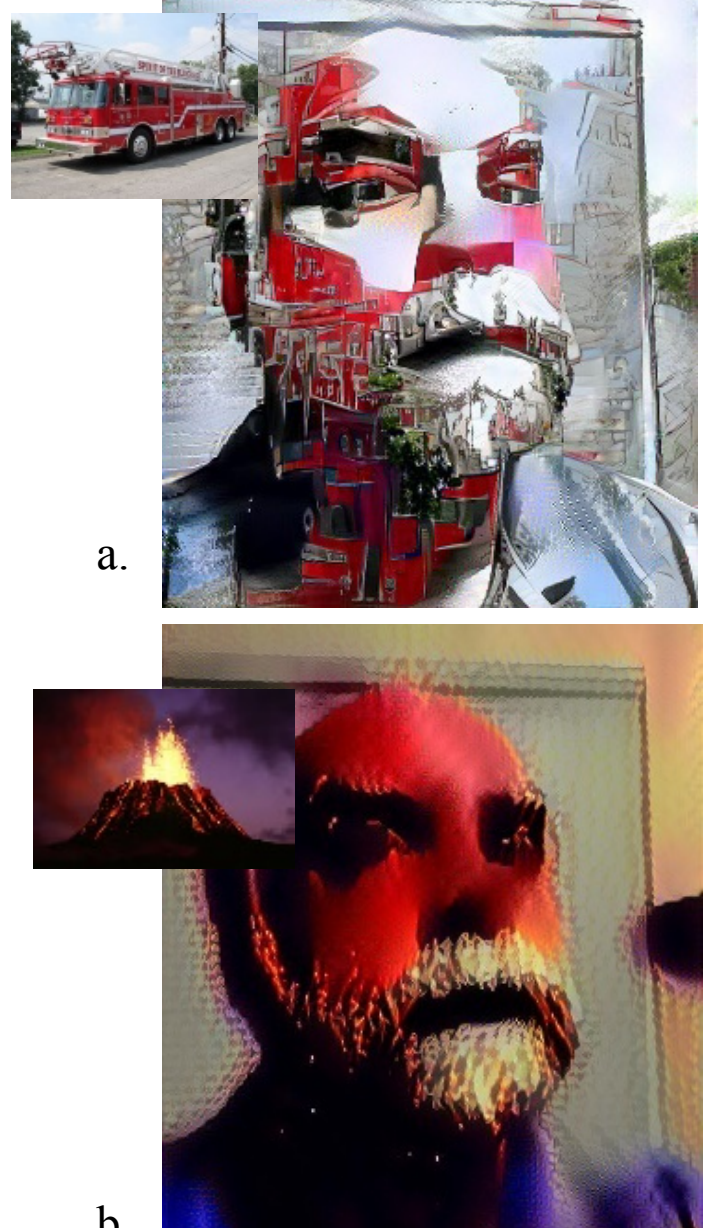

Figure 6. Deep Style using face content and engine, volcano style (a, b).

As part of our first extension of the DD and DS algorithms, we aim to situate DD/DS within the wider scope of an artificial painter system. In this type of system, DD/DS or similar systems can play the role of the artist's perception and imagination (for abstraction, narrative, and emphasis), while a further artistic painting algorithm models the phase that occurs in manifesting imagery onto the canvas and dialoging with the artist's materials. Eventually it would make sense to have a cyclical interaction between the perception/imagination phase and the stroke-painting phase. In this first attempt, we apply the DD/DS module to the source photo first, followed by a painting phase.

Figure 9 shows a DD image before and after treatment with the ePainterly cognitively inspired painting system (in digital format, zoom in to appreciate detail). This ePainterly system, an extension to cognitive painting system, Painterly (DiPaola 2009), models the cognitive processes of artists using algorithmic, particle system and noise modules to generate artistic color palettes, stroking and style techniques. It is in the stroke-based rendering subclass of non-photorealistic rendering (NPR) and is used as the final part of the process to realize the internal DD/DS models. In this example, aesthetic advantages include reducing some artifacting of DD output via cohesive stroke-based clustering as well as a better distributed color space.

## Evaluating Novelty and Aesthetic Value

We can also analyze DD and DS in terms of other ideas about computational creativity. We will examine to what extent these algorithms are able to pursue generated outputs having novelty and value (widely accepted as the defining characteristics of creative production).

We first inquire whether DD/DS make explicit autonomous evaluations of the novelty and value of their outputs. This question can be viewed as important for creativity: for example, Jennings (2010) includes such autonomous evaluation as a necessary condition for creative autonomy. (He also includes the ability to *change* evaluative standards, a condition not met by DD/DS on their own). In terms of novelty, when used as standalone systems, neither DD nor DS maintain or optimize explicit measures of novelty as an image is produced. In terms of value, the situation is more favorable: DD and DS isolate and maximize subsets of the network features evoked by each input image, resulting in the maintenance or enhancement of certain aspect image qualities (or whole levels of abstraction) at the expense of other qualities. This process can be construed as the computation of one or more aesthetic value metrics. It bears a resemblance to neuroaesthetic principles of art, such as Zeki's (2001) notions of highlighting/stimulating discrete portions of visual processing and translating the brain's abstractions on to the canvas or Ramachandran and Hirstein's (1999) laws of peak-shift and isolation.

Going beyond the ability of DD/DS to explicitly evaluate and optimize novelty and value, we can further ask whether DD/DS at least tend to generate outputs with high novelty and value. Informally, we do observe that the range of outputs generated by these systems often strike us as surprising while also being visually pleasing. The viewer may be surprised not only by the form of an unexpected visual concept being "imagined" into the original image (in the case of free hallucination with DD) but in the form of an unexpected manner of visual resemblance between selected inputs (e.g. Figure 7). Perhaps the network's support of a large number of multi-level visual features creates a rich space of combinations such that the optima found by DD/DS are not the same as the human viewer anticipates.

In Ritchie's (2007) empirical framework for assessing computational creativity, it is suggested that measures more fine-grained or primitive than novelty are dissimilarity from the "inspiring set" and typicality (relative to the target art form or genre). For DD and DS, we take the inspiring set to include both the original neural network training image set and the current input image(s) to the algorithm. It seems obvious that DD/DS are not simply replicating or near-replicating any of the training set images. By design, DD and DS outputs do bear visual similarity to the input images, but our general impression is that the resemblance is not so slavish as to exclude creativity. We furthermore observe a tendency for DD to find a more abstract or remote similarity with the style-guide image compared to DS, which aligns well with the fact that DS uses a more informationally rich optimization goal for style.

Ritchie regards typicality as a double-edged sword: on one hand, it is an achievement for a computer to generate successful examples of a style, but on the other hand, high typicality suggests low novelty. For DD, the comparison set for typicality is not obvious ("contemporary art" would be one choice), while DS will tend to be compared to the artist or genre of the style-guide image. Thus, to the extent that DS seems often to do a good job of mimicking an artist's style (e.g. Figure 8), Ritchie's approach might lead us to characterize it as impressive by one standard yet prevented from being creative at the highest level. Additional considerations from Ritchie (2007) regarding repetition point to other limitations on DD/DS's level of creativity. Repetition in the output arises when using DD/DS multiple times with the same inputs and parameter settings.

Finally, it is worth considering that DD/DS algorithms may be particularly amenable to future extensions and modifications that would enhance their ability to internally search for novelty/value. Regarding novelty, the vector spaces formed by node activations can lead to distance measures that are closer to human-perceived visual similarity compared to raw-pixel-based measures. Such measures could be combined with storage of images and data clustering techniques to estimate the novelty of a particular generated image compared to training images or compared to a certain corpus (e.g., art of a certain genre). Regarding aesthetic value, ideas from information-based aesthetic theories (Rigau et al. 2008) such as compressibility could be applied as additional optimization constraints, using the node vectors as a basis.

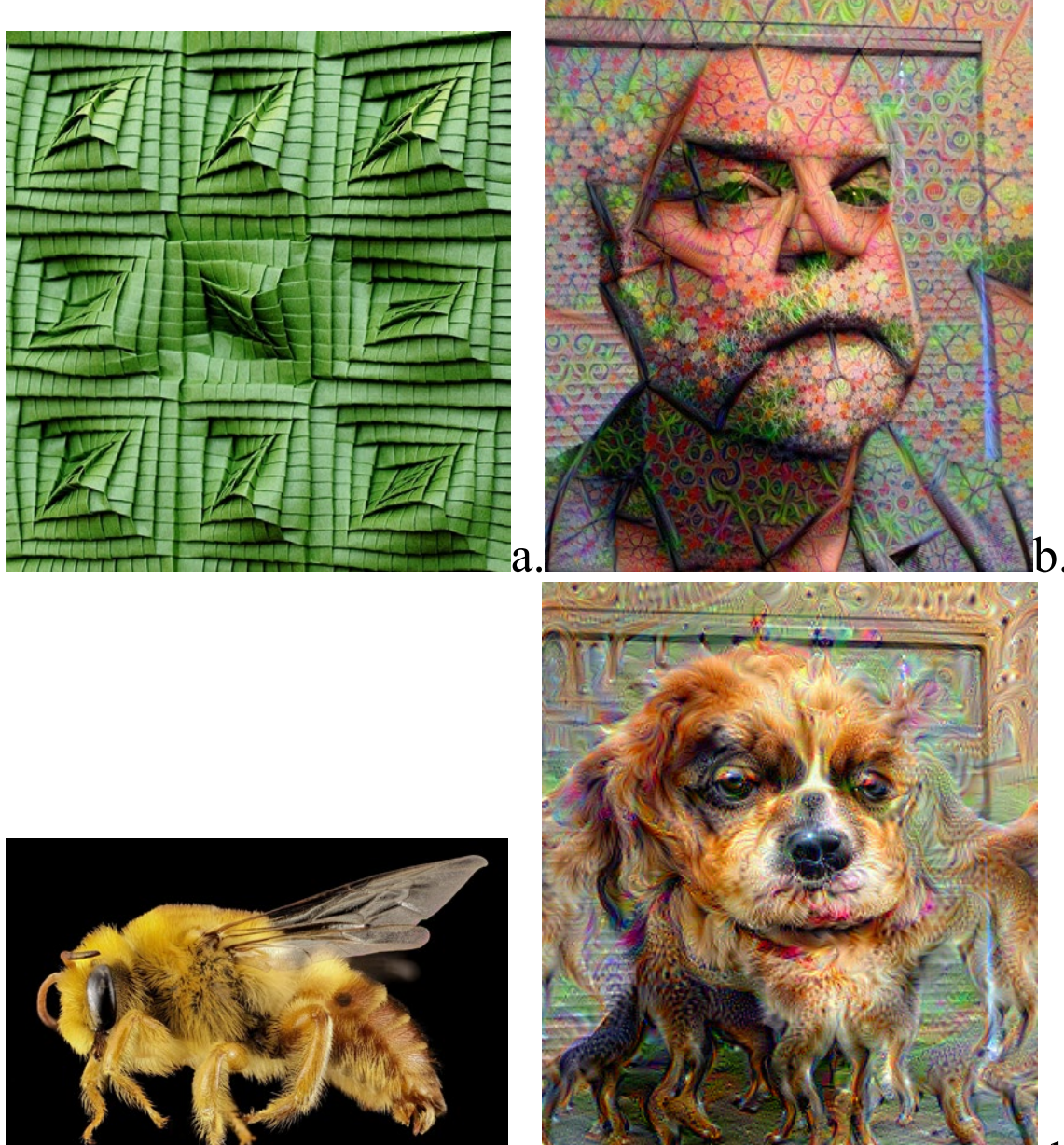

Figure 7. Deep Dream—surprising interpretations of image attributes by network. Image (a) by Goran Konjevod (Flickr, CC BY-NC 2.0)

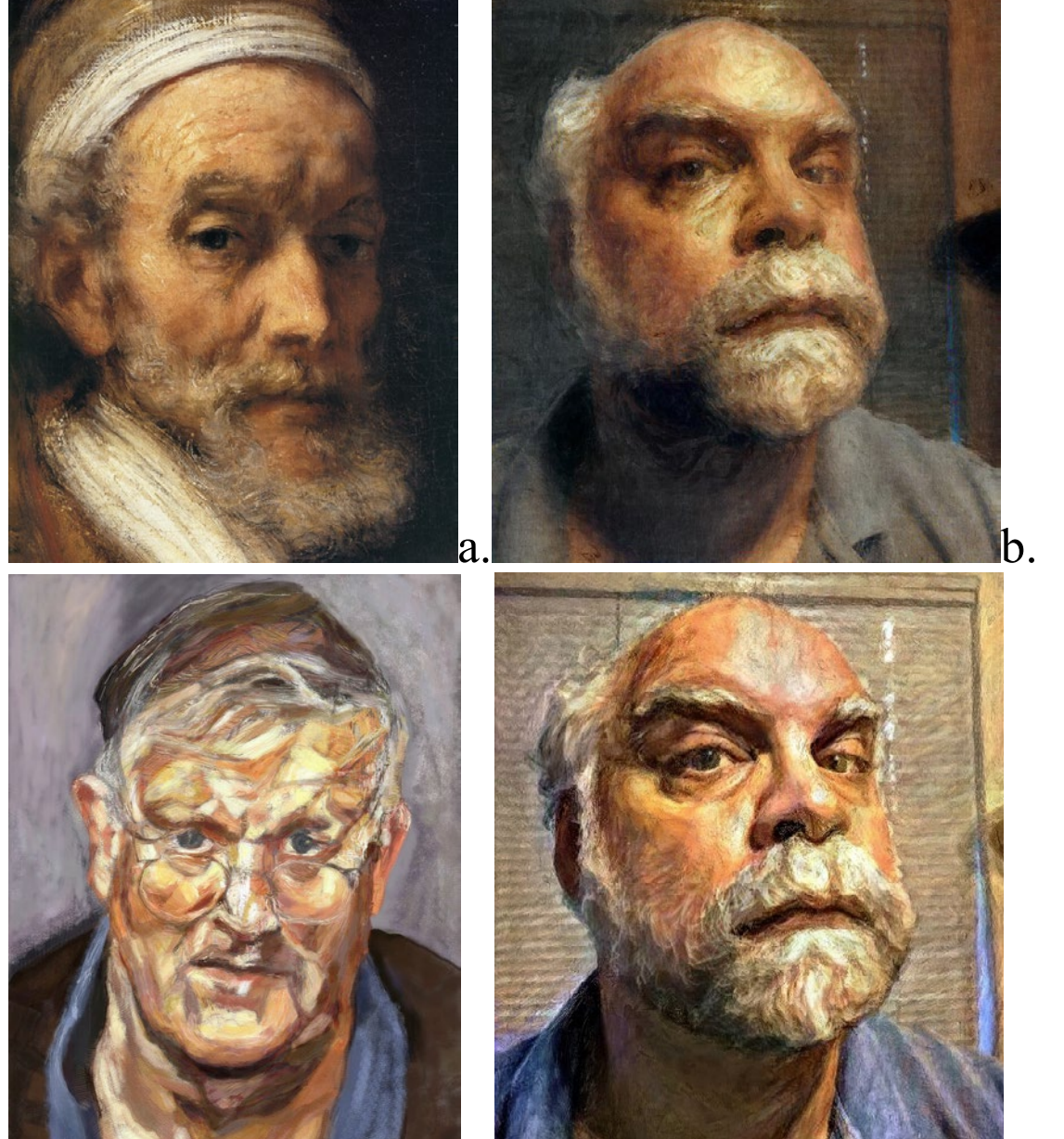

Figure 8. Deep Style using Rembrandt (a,b) and Freud (c,d) paintings as style guides.

## Conclusion and Future Directions

We suggest that the use of deep neural networks (with accompanying search/optimization algorithms) to produce creative visual blends and artifacts has entered a new and promising phase, both for models of creativity and for practical art-generating systems. Next steps include:

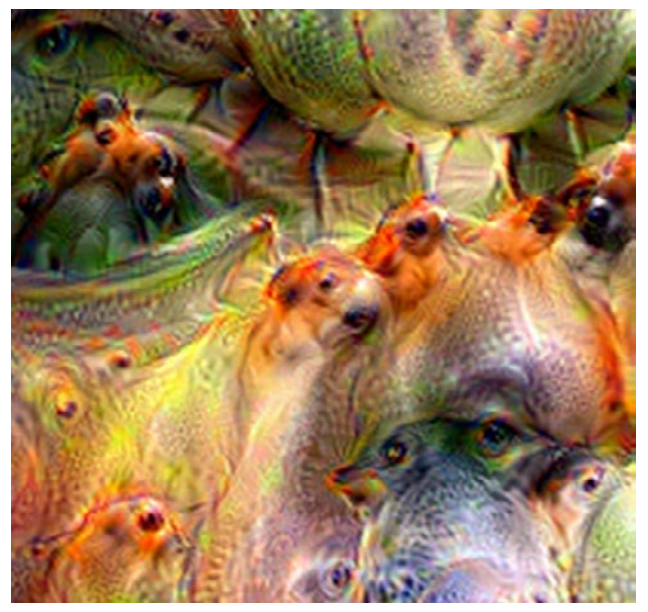
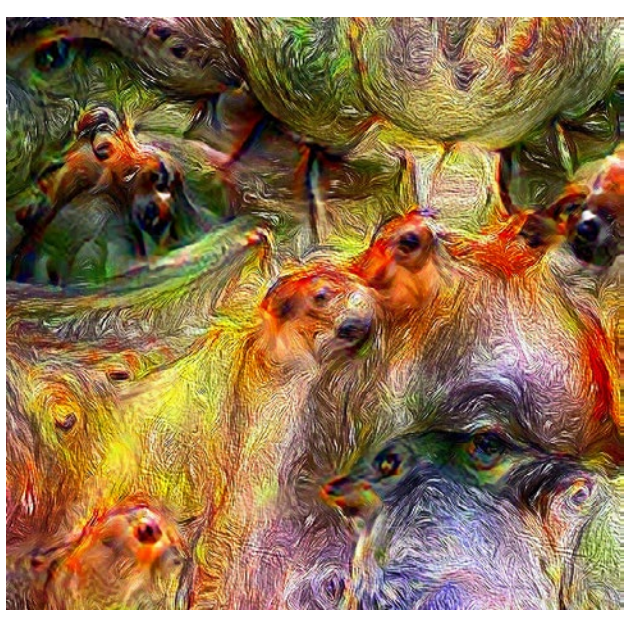

Figure 9. Deep Dream example (zoomed-in crop), before (l) and after (r) cognitive stroke-painting simulation.

1. Addressing flaws and limitations in current network architectures and training sets—some current work aims to increase network ability to generate more realistic images, avoiding the current visible artifacts (Denton et al. 2015). Richer or purpose-built training data should help get around issues of bias such as the "dog faces everywhere" effect.
2. Embedding deep neural nets as perceptual modules in larger architectures for cognitive creativity and art—other modules to add include emotional expression and planning/interaction with art materials. Examples of this direction include (Stewart et al. 2012; Augello et al. 2013).
3. Linking the CNN operation to different training paradigms or text-understanding modules—this could allow the networks to more explicitly represent concepts beyond noun-like object semantics and parts, for example allowing image generation based directly on adjectives at a visual or emotional level (e.g. "jagged", "joyful", or "bold").
4. Exploring augmentations to DD, DS for novelty and aesthetic evaluation as mentioned in the previous section.

## Acknowledgements

This work was supported by the National Sciences and Engineering Research Council of Canada (NSERC). Thanks to Daniel McVeigh and Jonathan Waldie for their assistance generating image examples. Our DS implementation is based on: https://github.com/fzliu/style-transfer .